\documentclass[letterpaper, 10 pt, conference]{ieeeconf}  

\IEEEoverridecommandlockouts                              

\usepackage{graphics} 
\usepackage{epsfig} 
\usepackage{mathptmx} 
\usepackage{times} 
\usepackage{amsmath} 
\usepackage{amssymb}  
\usepackage{color,soul}
\usepackage{url}

\usepackage{subfigure}
\usepackage{pgf}
\usepackage{booktabs}
\usepackage{dblfloatfix}
\usepackage{capt-of}
\usepackage{cuted}

\title{\LARGE \bf
From Prompts to Protocols: An AI Agent for Laboratory Automation
}

\author{Angelos Angelopoulos$^{1}$, James F. Cahoon$^{2}$, Ron Alterovitz$^{1}$%
\thanks{*This work was supported by the Creativity Hub at UNC-Chapel Hill.}%
\thanks{$^{1}$Angelos Angelopoulos and Ron Alterovitz are with the Department of Computer Science, University of North Carolina at Chapel Hill, NC 27599, USA
	{\tt\small \{aangelos,ron\}@cs.unc.edu}}%
\thanks{$^{2}$James Cahoon is with the Department of Chemistry, University of North Carolina at Chapel Hill, NC 27599, USA
	{\tt\small jfcahoon@unc.edu}}%
}%

\begin{document}

\maketitle
\thispagestyle{empty}
\pagestyle{empty}


\begin{abstract}
	
Automating science laboratories enables faster, safer, more accurate, and more reproducible execution of protocols, accelerating the discovery and testing of new materials, drugs, and more. However, setting up and running autonomous labs requires coordinating numerous instruments and robots, forcing scientists to write code, manage configuration files, and navigate complex software infrastructure. We present an AI agent architecture that integrates large language models with laboratory orchestration, enabling scientists to interactively create and monitor automated lab protocols using natural language. Integrated into the Experiment Orchestration System (EOS), the AI agent operates under an agentic loop with automated validation and error correction, and supports the complete experimental lifecycle: creating protocols, running and monitoring both protocols and closed-loop optimization campaigns, and analyzing results. A visual graph editor renders protocols as interactive node-based diagrams synchronized with the AI agent's protocol representation, enabling seamless alternation between AI-assisted and manual protocol construction. Evaluated on three simulated automated labs spanning chemistry, biology, and materials science, the AI agent achieves a 97\% first-attempt protocol generation success rate and an order of magnitude reduction in required interface actions.
	
\end{abstract}


\section{Introduction}

Laboratory automation can accelerate the discovery and testing of new materials, drugs, and other products by enabling scientists to run protocols faster, more safely, more accurately, and with greater reproducibility~\cite{Angelopoulos2024_Transforming,Abolhasani2023_Rise, Sanderson2019_Automation}. Lab automation can range from automating a single instrument to coordinating complex multi-step protocols across distributed laboratories~\cite{Angelopoulos2024_Transforming}. Setting up and running autonomous labs typically requires coordinating numerous instruments and robots, which must be scheduled and operated to complete experiments. To support this coordination, autonomous labs often use laboratory orchestration software that provides scientists with standard interfaces for defining and executing multi-step protocols. Orchestrators can automatically schedule when and on what instruments tasks run, manage connectivity to diverse lab hardware, collect and organize data from protocol runs, and provide frameworks to optimize protocol parameters~\cite{Angelopoulos2025_Experimenta, Fei2024_AlabOSa, Zhang2025_IvoryOSa}. However, creating and monitoring protocols using laboratory orchestration software remains tedious. Scientists must often implement code using Python or visual block-based programming, author configuration files, understand scheduling semantics, and monitor execution across multiple subsystems. These tedious steps, for which chemists and other physical scientists currently receive little training, serve as a barrier to adoption of lab automation~\cite{Angelopoulos2024_Transforming}.

\begin{figure}[t]
    \vspace{0.7em}
	\centering
	\includegraphics[width=\columnwidth]{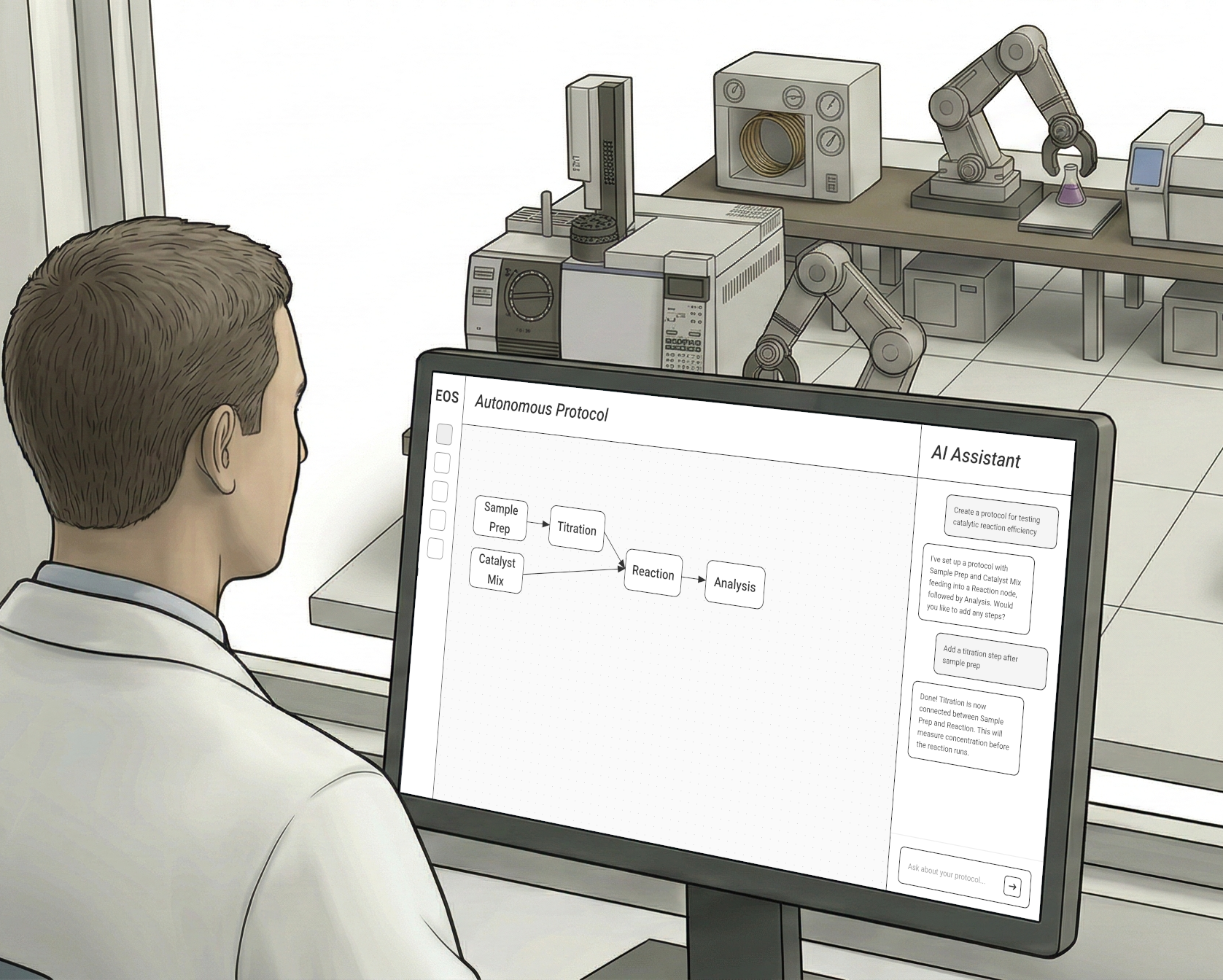}
	\caption{The EOS AI agent enables scientists to create experiment protocols, monitor experiment campaigns, and analyze results through natural language, coupled with an interactive visual graph editor.}
	\label{fig:header}
\end{figure}

In this paper, we present an AI agent architecture that integrates large language models with laboratory orchestration, enabling scientists to interactively create and monitor automated protocols using natural language. The AI agent, powered by state-of-the-art large language models, enables scientists to create protocols, submit protocols for execution, monitor protocol runs, run optimization campaigns with these protocols, and query and analyze generated data, all without writing code or configuration files. For example, a scientist can ask the AI agent to ``add 10mL of each of the three input reagents into the flask, mix using a magnetic mixer for 20s, and characterize the solution using mass spectrometry'', and the AI agent generates a complete, validated protocol (represented by a directed acyclic graph of tasks and associated parameters) by reasoning about available devices, implemented task capabilities, and laboratory constraints. 

We integrated this AI agent into the Experiment Orchestration System (EOS), a laboratory orchestration software supporting protocol creation and monitoring~\cite{Angelopoulos2025_Experimenta}. The EOS AI agent operates under a full agentic loop: it reasons about the scientist's request, calls Model Context Protocol (MCP) tools such as to read task specification files to gather information and take actions, validates its generated protocol and other outputs using the EOS orchestrator's validation engine via MCP to check structural correctness (e.g., check that parameter values are within bounds), and automatically detects and corrects errors until its task is complete. We implemented the integration of the AI agent with EOS by exposing over 40 tools through an MCP server that covers nearly all EOS functionality. The AI agent supports the complete experimental lifecycle, from creating protocols in EOS, to running and monitoring protocols and closed-loop optimization campaigns, to analyzing results. 

We coupled the EOS AI agent with a visual graph editor that renders protocols as interactive node-based diagrams. This allows scientists to visually verify and refine the AI-generated protocols. Users can also edit the diagrams in a GUI if desired. These diagrams are synchronized in real-time with the AI agent's internal representation of the protocol, enabling users to seamlessly alternate between AI-assisted protocol creation and manual GUI-based protocol creation. 

In this paper, we describe how the EOS AI agent enhances the experiment lifecycle from the scientist's perspective in Section~\ref{sec:ai}, detail the implementation in Section~\ref{sec:implementation}, and demonstrate the AI agent in Section~\ref{sec:demonstration} on three representative simulated lab protocols: color mixing, solubility and purification screening, and a liquid-liquid extraction.


\section{Related Work}

Automated experimentation coupled with machine learning can create self-driving labs that accelerate discoveries in the physical and life sciences~\cite{Abolhasani2023_Rise,Sanderson2019_Automation,Angelopoulos2024_Transforming}. Autonomous platforms have synthesized novel materials~\cite{Szymanski2023_Autonomous}, discovered molecular candidates meeting multiple property targets~\cite{Koscher2023_Autonomous}, and navigated chemical reaction spaces with machine learning~\cite{Granda2018_Controlling}. Physical automation has progressed from single-purpose apparatuses to mobile robots that transport samples between stations~\cite{Burger2020_Mobile,Angelopoulos2023_HighAccuracy} and modular flow synthesis systems guided by AI~\cite{Coley2019_Robotic} or domain-specific languages~\cite{Steiner2019_Organic, Mehr2020_Universal}. Bayesian optimization is widely used to direct closed-loop campaigns for autonomous process optimization~\cite{Christensen2021_Datascience} and rapid materials discovery~\cite{Kusne2020_Onthefly}. As these labs grow in complexity, orchestration and usability become critical bottlenecks~\cite{Angelopoulos2024_Transforming}.

On the software side, early lab orchestration systems introduced scheduling and parallelism for automated workstations~\cite{AndrewCorkan1992_Experiment}, while recent orchestrators have focused on coordinating heterogeneous instruments within self-driving labs~\cite{Roch2020_ChemOS, Angelopoulos2025_Experimenta}. Visual programming environments have sought to make protocol creation accessible to non-programmers~\cite{Gupta2017_BioBlocks}, and operating-system-like abstractions have been proposed to unify lab software infrastructure~\cite{Segal2019_Operating}. Large language models (LLMs) have been applied directly to lab settings: Coscientist used GPT-4 with tool-calling capabilities to autonomously plan and carry out catalytic cross-coupling optimizations~\cite{Boiko2023_Autonomous}, ChemCrow augmented an LLM with eighteen specialized chemistry tools spanning synthesis planning, safety assessment, and property prediction~\cite{M.Bran2024_Augmenting}, and ORGANA combined an LLM-based reasoning layer with task planning~\cite{Darvish2025_ORGANAa}. However, these LLM-based systems typically function as standalone agents that interact with individual instruments or computational backends rather than with a full-featured lab orchestrator.

IvoryOS~\cite{Zhang2025_IvoryOSa} is a user-friendly Python-based orchestrator that dynamically generates web interfaces to control hardware by introspecting Python scripts. It includes an LLM-based natural language mode for protocol generation. However, the integration is stateless and single-turn, with no conversation history, tool calling, or iterative refinement. Validation errors are not fed back to the LLM for correction, the generated protocols are sequential action lists without support for task dependencies or parallel branches, and the LLM's scope is limited to protocol creation. In contrast, the EOS AI agent offers a full agentic loop that reasons across multiple steps, generates protocols as directed acyclic graphs with parallel branches, feeds validation errors back to the LLM for automatic correction, and extends beyond protocol creation to monitoring and analysis.

AlabOS~\cite{Fei2024_AlabOSa} is a Python framework for autonomous lab protocols, demonstrated at scale in the A-Lab facility where it orchestrated the synthesis of over 3500 material samples. The system employs a manager-worker architecture with centralized resource management and defines protocols as directed acyclic graphs. AlabOS provides no AI assistance or natural language interface; scientists must author protocols using Python code. The EOS AI agent enables scientists to create, submit, and monitor running protocols through conversation.

ChemOS 2.0~\cite{Sim2024_ChemOS} is a modular orchestrator where components communicate via SiLA2~\cite{Juchli2022_SiLA}, integrating computational simulations alongside physical protocols for closed-loop campaigns. The system provides a Streamlit-based dashboard where scientists submit jobs by uploading JSON parameter files and retrieve results by querying job identifiers. However, ChemOS 2.0 offers no natural language interface; multi-step protocols require Python code to chain SiLA2 device calls and coordinate devices via threading. In contrast, the EOS AI agent allows scientists to use natural language to generate complete protocols, query results, and manage automated labs.


\section{Enhancing Laboratory Automation with AI}
\label{sec:ai}

\begin{figure*}[t]
    \vspace{0.5em}
	\centering
	\includegraphics[width=\textwidth]{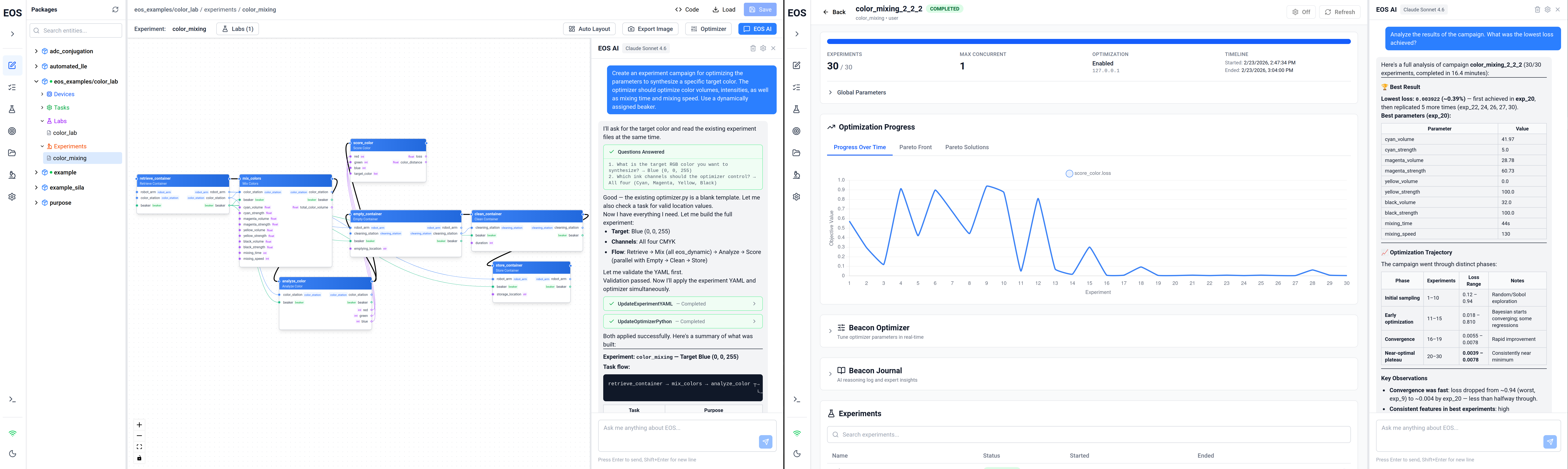}
	\caption{The EOS AI agent is integrated with the EOS user interface. \textbf{Left}: the protocol editor with the AI agent creating a protocol. \textbf{Right}: optimization campaign progress tracking with AI-assisted result analysis.}
	\label{fig:ui_overview}
\end{figure*}

The EOS AI agent supports scientists throughout the experimental process, from creating protocols to monitoring execution and analyzing results. We integrated the AI agent into the EOS web interface so it is available across all views, as shown in Figure~\ref{fig:ui_overview}. It adapts its behavior to the scientist's current context within the interface. In the following subsections, we describe the AI agent's capabilities from the scientist's perspective.

\subsection{Creating Protocols}

The EOS AI agent can assist scientists in creating protocols, a fundamental task in lab automation. Scientists describe their goals in natural language, and the AI agent generates complete protocols formatted as directed acyclic graphs of tasks with parameters. To accomplish this, the AI agent receives as context from EOS the full set of available laboratory resources, including task specifications, hardware device specifications, resource constraints, as well as the current state of the protocol being created. This context enables the AI agent to create protocols that reference existing devices, respect parameter constraints, and satisfy task dependencies.

As the AI agent generates or modifies a protocol, changes appear in real time in a visual graph editor that visually renders the protocol. Task nodes appear on the canvas, directed edges form between dependent tasks, and parameter configurations update as the AI agent works. If scientists want to manually intervene, they can edit the visual graph directly, and their changes are preserved in the same underlying protocol representation that the AI agent operates on, enabling seamless alternation between AI-assisted and manual protocol construction.

 The AI agent automatically validates its generated protocols against the EOS orchestrator and iterates to correct errors without requiring the scientist to intervene. EOS has an extensive validation engine that covers both structural correctness (e.g., referenced devices exist and dependencies are acyclic) and scientific constraints defined in task specifications (e.g., parameter ranges and units). All errors are batched and returned together, enabling the AI agent to address every error in a single correction step.

\subsection{Monitoring Protocol Runs and Campaigns}

Beyond protocol creation, the AI agent assists scientists in monitoring protocol runs and optimization campaigns. Scientists can query the AI agent about the status of tasks, protocol runs, and campaigns through natural language rather than navigating multiple data views and interpreting raw information. For individual protocol runs, the AI agent can report which tasks have completed, which are currently running, and which are pending, contextualizing this information within the protocol's dependency structure to identify bottlenecks or failures that may be blocking downstream execution. For optimization campaigns, the AI agent can explain optimization progress, convergence behavior, and explored parameter regions. The EOS interface also provides visual monitoring tools such as objective value charts and Pareto front visualizations, which scientists can discuss with the AI agent.

\subsection{Analyzing Experimental Data}\label{sec:analyzing}

The AI agent can directly query EOS's database containing data from protocol runs by executing SQL queries as well as executing code to analyze the data. Scientists can ask the AI agent to retrieve results, compute summary statistics, identify trends, or compare outcomes across protocol runs. Beyond experimental results, the agent can access lab and device definitions, inspect live device states, and browse task and protocol source files. Rather than exporting data to external analysis tools, scientists can analyze their results conversationally. To ensure data integrity, we have given the EOS AI agent read-only access to the database.

\subsection{Submitting Work and Managing the Laboratory}

The AI agent can submit individual tasks, protocols, and optimization campaigns for execution on behalf of the scientist through natural language instructions, eliminating the need to manually run programs or navigate a GUI. The AI agent can also perform laboratory administration and interact directly with hardware, such as querying device state or invoking device functions. These capabilities are especially useful during development and debugging, when scientists need to verify device behavior or iteratively test new task code. To maintain safety, all mutating operations require explicit user approval, while read-only operations execute automatically.


\section{Implementation}
\label{sec:implementation}

Our implementation of the EOS AI agent includes a front-end that enables the user to interact with the AI agent and a dedicated backend that handles the AI agent’s core logic and serves as the bridge to external systems. Specifically, the front-end can be embedded in a web application and the backend communicates with the EOS orchestrator via a REST API and has read-only access to the orchestrator’s PostgreSQL database for data queries. This architectural separation allows the AI agent and its interface to be developed and deployed independently of the EOS orchestrator. The EOS AI agent currently integrates with Claude Sonnet 4.6 and Claude Opus 4.6, and support for additional models will be added.

\subsection{AI Agent Architecture}

\begin{figure}[t]
	\centering
	\includegraphics[width=\linewidth]{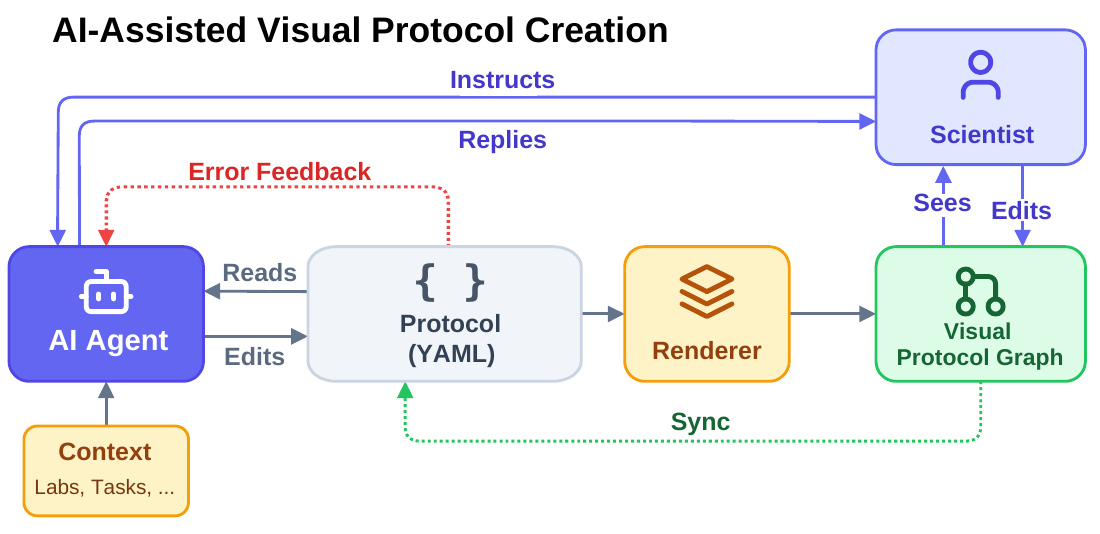}
	\caption{Architecture of AI-assisted protocol creation. The AI agent uses laboratory context to edit a protocol defined in YAML (structured text), which a renderer converts into a visual graph. The scientist can instruct the AI agent or edit the graph directly, with changes synced bidirectionally. Validation errors feed back to the agent for automatic correction.}
	\label{fig:ai_assistant_diagram}
	\vspace{-0.5em}
\end{figure}

Figure~\ref{fig:ai_assistant_diagram} illustrates the architecture of the AI-assisted protocol creation. The EOS AI agent operates under a full agentic loop that supports multi-turn conversations with persistent session state. When a scientist sends a natural language prompt, the AI agent prepends a dynamically generated system prompt that provides the backend model with comprehensive laboratory context. This system prompt includes the EOS domain model, task specifications with their parameter schemas and device requirements, device specifications from available laboratories, and descriptions of the available MCP tools. The system prompt also has dynamic elements that change based on the front-end view the scientist is looking at. For example, if they are looking at the campaign progress view, the AI agent receives information about the campaign, such as its name.

The AI agent's protocol creation process proceeds as follows. The agent calls tools to acquire additional context, such as reading existing protocols or querying available device specifications, and then edits the protocol, which is defined using YAML, a structured text representation. A separate validation tool sends the protocol to the EOS orchestrator for validation, as described in Section~\ref{sec:ai}. A batched error logger collects all errors in text format and returns them in a single response. The complete set of errors is fed back to the agent, which iterates and re-validates until the protocol is valid or a maximum number of reasoning steps is reached. The validated protocol is then parsed by a renderer that converts it into the visual protocol graph. Because the visual protocol graph and the protocol YAML share a single underlying data store, scientists can also edit the graph directly, and their changes propagate back to the protocol YAML.

Beyond protocol creation, the AI agent has access to tools spanning nearly all of EOS's functionality, as described in Section~\ref{sec:MCP}. We also equipped the AI agent with a question-asking tool so it can seek feedback from the scientist when it encounters underspecified or ambiguous situations. The AI agent can ask up to 10 questions, each with several possible answer choices generated by the AI agent as well as a field for a custom answer if needed. The AI agent is free to use this tool whenever it wants, including to ask follow-up questions.

\subsection{Visual Protocol Rendering}

The visual protocol editor renders protocols as node-based directed acyclic graphs, visualizing both AI-generated and manually authored protocols. Task nodes expose color-coded typed ports for dependencies, devices, resources, and parameters, and the editor enforces connection validity by checking that linked ports share compatible types. Protocols are represented internally as a list of task nodes, each storing its type, canvas position, device and resource assignments, parameter values, and dependency list. This structure maps directly to the YAML-based protocol definition that EOS consumes, so any EOS protocol defined in YAML can be rendered in the visual protocol graph editor and vice versa. A centralized state store serves as the single source of truth since both the AI agent and user's direct edits write to this store, ensuring consistency. When the AI agent modifies a protocol, a collision avoidance postprocessor shifts nodes apart to prevent visual overlaps.

\subsection{Model Context Protocol (MCP) Server}
\label{sec:MCP}

\begin{figure}[t]
    \vspace{0.5em}
	\centering
	\includegraphics[width=\linewidth]{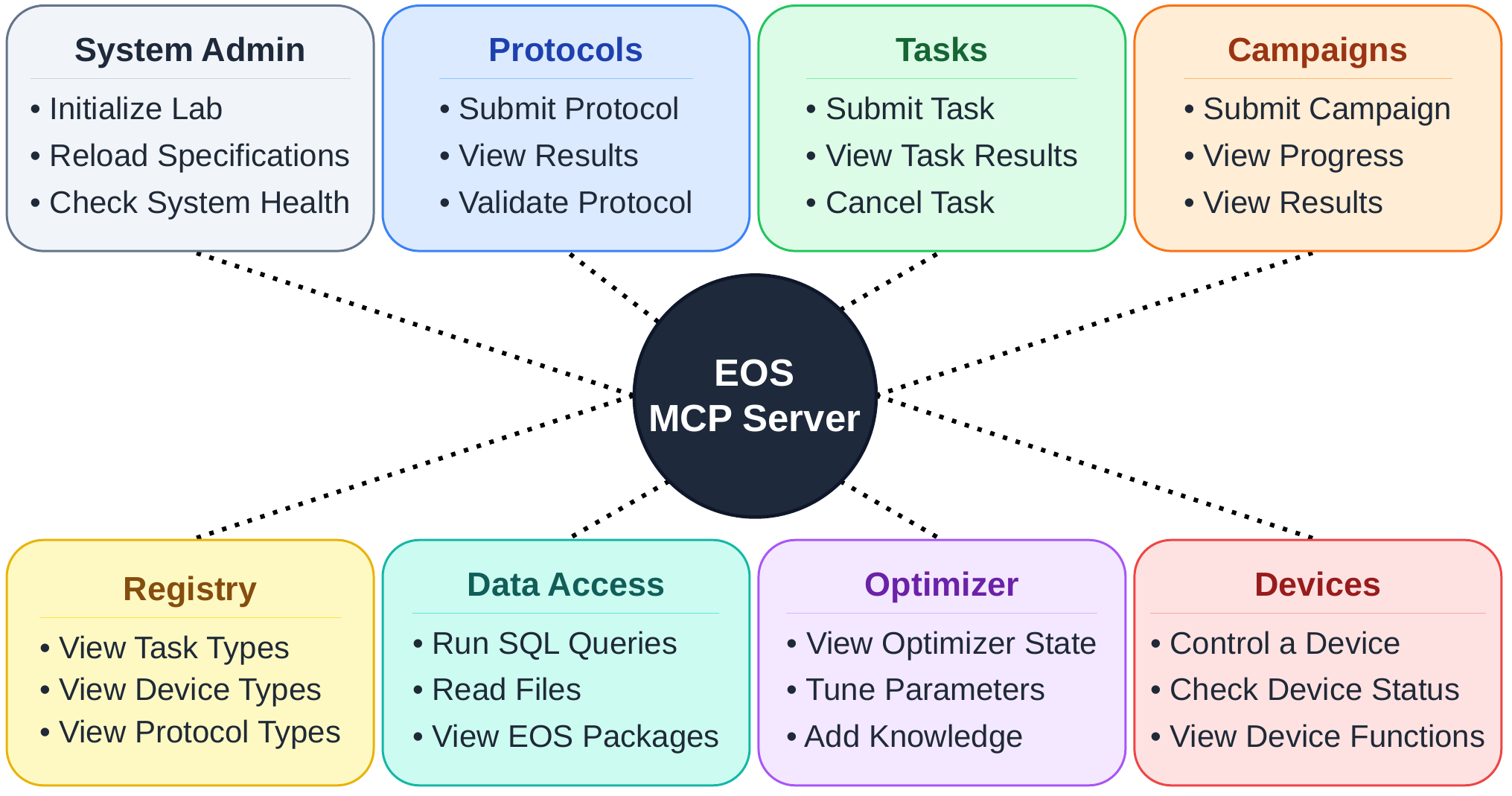}
	\caption{The EOS MCP server organizes over 40 tools into eight categories. Some representative tools per category are shown.}
	\label{fig:eos_mcp_tools}
\end{figure}

The backend of the EOS AI agent runs a Model Context Protocol (MCP) server that allows both the EOS AI agent and AI agents outside of the EOS user interface to interact with EOS through a standardized tool interface. The MCP server runs as an in-memory bridge within the backend process, avoiding network round-trips for the integrated agent while still exposing the standard MCP protocol for external clients. As shown in Figure~\ref{fig:eos_mcp_tools}, the server organizes over 40 tools into functional categories: tasks, protocols, and campaigns; devices; system administration; optimizer; data access; and registry. The tools are divided into two classes based on their safety implications. Read-only tools execute automatically without interrupting the AI agent's reasoning, while mutating operations require explicit user approval before execution.


\section{Demonstration}\label{sec:demonstration}

We evaluate the EOS AI agent on three simulated scenarios introduced in prior work: creating and executing a closed-loop color mixing optimization campaign in a virtual laboratory~\cite{Angelopoulos2025_Experimenta}, creating a protocol for automated solubility and purification~\cite{2023_Hein}, and creating a protocol for automated liquid-liquid extraction~\cite{2023_Hein}. We also demonstrate the agent's ability to extract and analyze experimental data from a completed campaign.

\subsection{Color Mixing Optimization Campaign}

\begin{figure*}[!t]
	\centering
	\includegraphics[width=\textwidth]{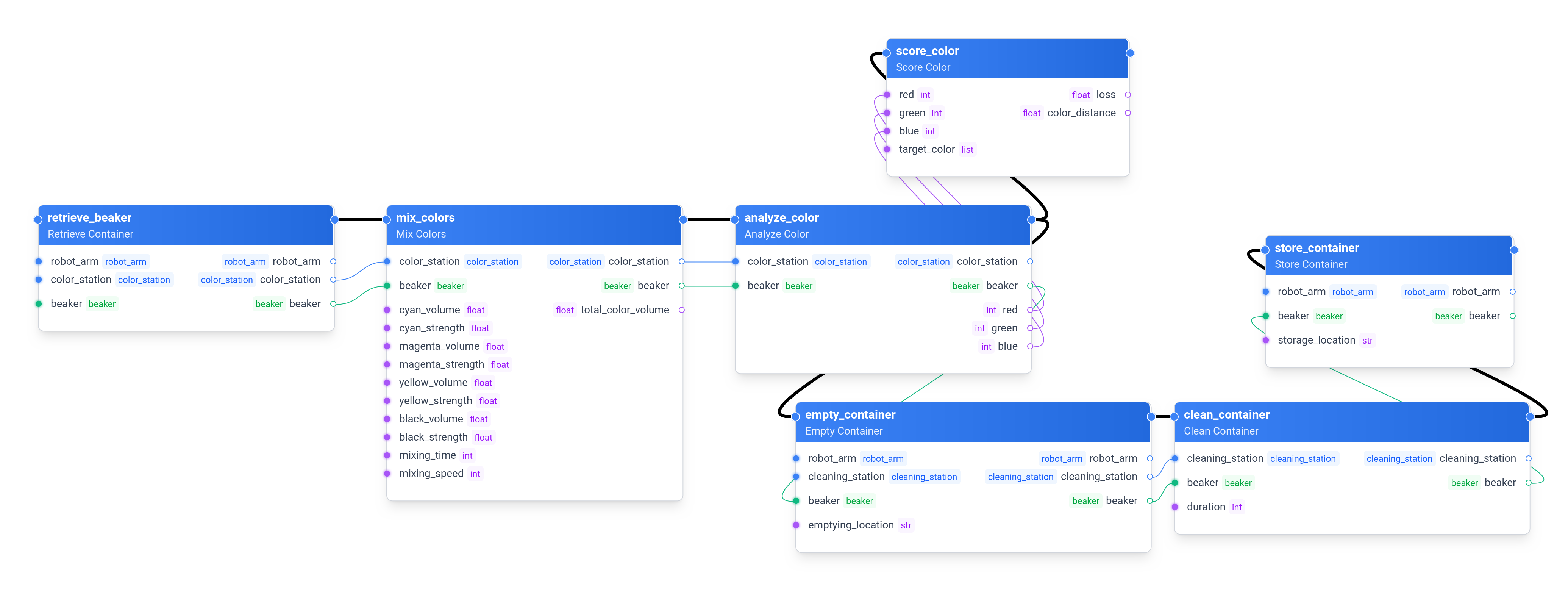}
	\caption{The color mixing protocol generated by the EOS AI agent.}
	\label{fig:color_mixing_workflow}
\end{figure*}

The objective of the color mixing optimization campaign is to synthesize a target RGB color from 4 ingredient colors (cyan, magenta, yellow, and black). There are 10 input parameters: volume and strength (how diluted the color is) for each of the four ingredient colors, plus mixing time and speed. The loss function is the Euclidean distance between the synthesized color and the target color. The protocol runs in a virtual laboratory with the following devices: a robot arm, three color dispensing and detection stations, and a container cleaning station. The virtual lab simulates color mixing using a real-time GPU-accelerated fluid solver~\cite{Dobryakov2024_PavelDoGreat} based on Stable Fluids~\cite{Stam1999_Stable}.

To evaluate the performance of the AI agent, we gave it the following prompt: ``Create an experiment campaign for optimizing the parameters to synthesize a specific target color. The optimizer should optimize color volumes, intensities, as well as mixing time and mixing speed. Use a dynamically assigned beaker.'' For this evaluation, the EOS AI agent leveraged Claude Sonnet 4.6. The EOS AI agent produced a complete protocol along with an optimizer definition in approximately three minutes. The resulting protocol graph is shown in Fig.~\ref{fig:color_mixing_workflow}.

To assess the EOS AI agent's reliability in creating protocols, we executed the same prompt 35 times and classified each outcome as either (a) correct, meaning the protocol and accompanying optimizer required no modifications, or (b) requiring correction via manual edits or follow-up prompts. The AI agent produced a correct protocol on the first attempt in 33 of 35 trials (94\% one-shot success rate). Of the two failures, one was a missing import statement in the optimizer Python code and the other was a logical error of skipping a task in the beginning of the protocol (mixing colors without retrieving a beaker first). We also recorded AI agent execution metrics shown in Table~\ref{tab:agent_metrics}.

To evaluate the EOS AI agent's ability to generalize across structurally diverse protocols, we tested three additional prompts that vary in desired graph topology, each executed 10 times. Table~\ref{tab:prompt_evaluation} summarizes the results. Prompt P1 tests mixing specific colors, P2 tests parallel branching with two independent sub-protocols, and P3 tests a compositional protocol where the output of one mix is used as input to the next. Across all four prompts (65 total trials), the agent achieved an overall first-attempt success rate of 97\%. 

\begin{table}[t]
    \vspace{0.5em}
	\caption{First-attempt protocol generation accuracy across different prompts.}
	\label{tab:prompt_evaluation}
	\centering
	\renewcommand{\arraystretch}{1.1}
	\resizebox{\columnwidth}{!}{%
		\begin{tabular}{@{}p{5.2cm}ccc@{}}
			\toprule
			\textbf{Prompt} & \textbf{Trials} & \textbf{Correct} & \textbf{Tasks} \\
			\midrule
			P0: Create a campaign for optimizing parameters to synthesize a target color. Optimize volumes, intensities, mixing time and speed. Use a dynamically assigned beaker. & 35 & 94\% & 7 \\
			\hline
			P1: Mix cyan and magenta to produce purple. Get a beaker from storage and put it back when done. & 10 & 100\% & 7 \\
			\hline
			P2: Create two separate color mixes in parallel, then compare their results. Get beakers from storage and put them back when done. & 10 & 100\% & 14 \\
			\hline
			P3: Mix colors in two stages: first mix cyan and magenta, then add yellow to the result. Get a beaker from storage and put it back when done. & 10 & 100\% & 4 \\
			\midrule
			\textbf{Overall} & \textbf{65} & \textbf{97\%} & -- \\
			\bottomrule
	\end{tabular}}
\end{table}

\begin{table}[t]
	\caption{AI agent execution metrics for color mixing protocol generation (N=35 trials).}
	\label{tab:agent_metrics}
	\centering
	\renewcommand{\arraystretch}{1.1}
	\begin{tabular}{@{}lcc@{}}
		\toprule
		\textbf{Metric} & \textbf{Mean $\pm$ Std} & \textbf{Range} \\
		\midrule
		Wall Time (s) & 184 $\pm$ 37 & 120--272 \\
		Reasoning Steps & 6.6 $\pm$ 2.8 & 3--14 \\
		MCP Tool Calls & 7.3 $\pm$ 3.9 & 3--16 \\
		Validation Corrections & 0.1 $\pm$ 0.2 & 0--1 \\
		LLM Cost (USD) & 0.50 $\pm$ 0.12 & 0.32--0.85 \\
		\bottomrule
	\end{tabular}
\end{table}

\begin{figure}[t]
    \vspace{0.5em}
	\centering
	\includegraphics[width=0.6\linewidth]{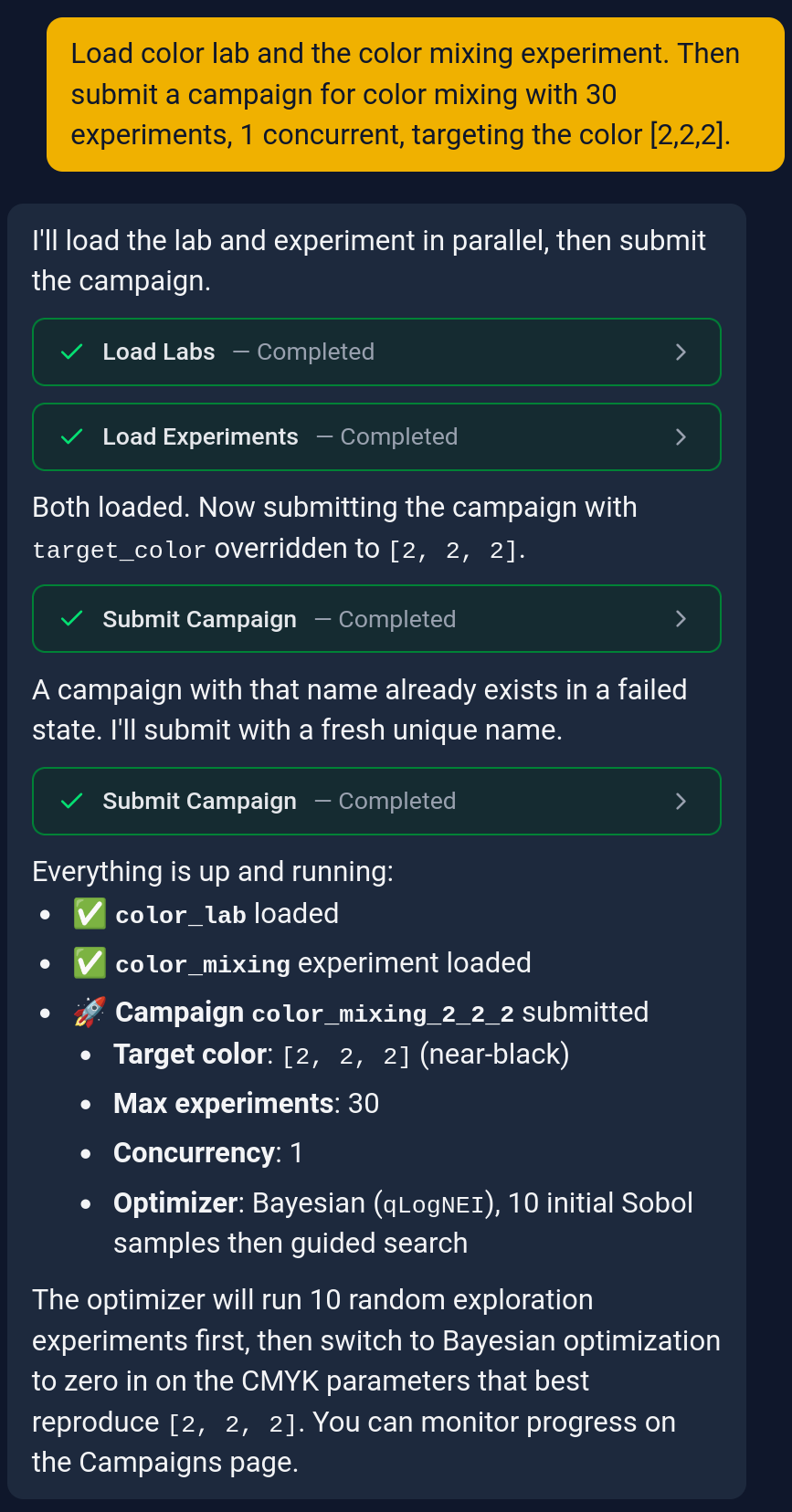}
	\caption{Conversation with the EOS AI agent to load the experiment protocol and associated laboratory code and submit a closed-loop optimization campaign.}
	\label{fig:ai_campaign_submission}
\end{figure}

After creating a protocol using prompt P0, we gave the EOS AI agent the prompt: ``Load color lab and the color mixing experiment protocol. Then submit a campaign for color mixing with 30 experiments, 1 concurrent, targeting the color [2,2,2].'' As shown in Fig.~\ref{fig:ai_campaign_submission}, the agent sequentially invoked MCP tools to load the lab, register the protocol, and submit a 30-experiment optimization campaign.
The campaign ran in the virtual lab, and the optimizer was able to synthesize the color [1,1,1], which is virtually identical to the target color.

To demonstrate the data analysis capability described in Section~\ref{sec:analyzing}, we queried the AI agent about the color mixing campaign results. When asked ``What was the best result in the campaign? What parameters produced it?'', the AI agent queried the EOS database and responded with the lowest loss, which was 0.00392 (approximately $1/255$, a near-perfect color match), and a table with the best values of the 10 input parameters. When asked ``How did the loss evolve over the campaign?'', the AI agent correctly summarized four convergence phases and then gave an accurate summary: ``The optimizer found near-optimal parameters by exp 16 (after just ~6 Bayesian iterations) and the campaign converged hard after exp 20. The remaining 10 experiments confirmed the optimum but didn't improve it further.'' Finally, when asked ``Which parameters had the most variation across the top~5 experiments?'', the AI agent wrote and executed analysis code, reported standard deviations of the 10 input parameters, and then stated the following insightful summary: ``Takeaway: yellow\_volume=0, yellow\_strength=100, and black\_strength=100 are locked constants in all top results --- these are the critical parameters. cyan\_strength, magenta\_strength, and magenta\_volume vary wildly, meaning the optimizer hasn't converged on them and they're likely low-sensitivity dimensions.''

\subsection{Protocol Creation for Solubility \& Purification Screening}

We also evaluated the ability of the EOS AI agent to create protocols for PurPOSE~\cite{2023_Hein}, a self-driving robotic chemistry platform developed at the University of British Columbia which served as a demonstration platform for IvoryOS~\cite{Zhang2025_IvoryOSa}. PurPOSE automates solubility screening and cooling crystallization purification of pharmaceutical compounds using 10 instruments including robot arms, a thermoshaker, an HPLC system, a liquid handler, and a centrifuge. The platform can execute a variety of processes including (1) HPLC standard curve calibration to establish concentration quantification, (2) solubility screening across solvent systems, and (3) cooling crystallization optimization with multi-objective Bayesian optimization over temperature difference, cooling rate, and final temperature to maximize crystal yield, purity, and impurity rejection. We extracted from PurPOSE's source code 10 device definitions and 16 task definitions to create a declarative specification (in YAML) defining the lab's devices and tasks for EOS, excluding any Python source code.

We asked the EOS AI agent to generate protocols for three PurPOSE processes using short natural language prompts. For the standard curve calibration process, the prompt was: ``Create experiment for standard curve for PurPOSE lab.'' For solubility screening process, the prompt was: ``Create experiment for solubility for PurPOSE lab.'' For the full crystallization campaign, the prompt was: ``Create EOS experiment campaign to optimize crystallization for PurPOSE lab. standard curve -$>$ solubility -$>$ crystallization.'' To create these protocols, the AI agent leveraged Claude Opus 4.6. The AI agent generated correct standard curve calibration (6 tasks) and solubility screening (5 tasks) protocols on its first attempt, producing task sequences structurally identical to those in the PurPOSE source code. For the crystallization campaign, which comprises 15 tasks across three phases together with a multi-objective optimizer, the agent's first attempt was a close reproduction but had some discrepancies: it specified a single optimization objective (yield) rather than three (yield, purity, and impurity rejection), and it made the shaking speed a variable optimization parameter when the PurPOSE implementation holds it constant.

These differences stem from the open-ended nature of the input prompt rather than from a limitation in the EOS AI agent. The agent made assumptions that required post-hoc correction. We also tested the agent with the question-asking tool described in Section~\ref{sec:implementation}, and the agent was able to identify ambiguous situations such as that the prompt did not specify optimization objectives and asked the scientist to clarify before generating the protocol. In several trials, the agent also proactively noted that the standard curve calibration phase is slow and asked if it should be separated from the end-to-end campaign so it does not need to be repeated in every run.

For all three processes, the EOS AI agent generated the correct task sequences and parameters without access to PurPOSE's underlying Python implementation code. This demonstrates that EOS's declarative YAML specification layer provides sufficient information for AI-driven protocol synthesis, and that the agent can handle complex, multi-phase protocols. Table~\ref{tab:agent_metrics_crystallization} reports agent execution metrics for the crystallization campaign across 5 trials.

\begin{table}[t]
    \vspace{0.5em}
	\caption{AI Agent execution metrics for PurPOSE crystallization campaign protocol generation (N=5 trials).}
	\label{tab:agent_metrics_crystallization}
	\centering
	\renewcommand{\arraystretch}{1.1}
	\begin{tabular}{@{}lcc@{}}
		\toprule
		\textbf{Metric} & \textbf{Mean $\pm$ Std} & \textbf{Range} \\
		\midrule
		Wall Time (s) & 410 $\pm$ 42 & 343--460 \\
		Reasoning Steps & 9.8 $\pm$ 2.7 & 6--12 \\
		MCP Tool Calls & 13.0 $\pm$ 2.9 & 9--16 \\
		Validation Corrections & 1.0 $\pm$ 1.0 & 0--2 \\
		LLM Cost (USD) & 1.10 $\pm$ 0.14 & 0.93--1.24 \\
		\bottomrule
	\end{tabular}
\end{table}

\subsection{Protocol Creation for Liquid-Liquid Extraction}

We also tested protocol synthesis for an automated liquid-liquid extraction (LLE) platform~\cite{Zhang2025_IvoryOSa}. The LLE platform integrates a UR3 robotic arm, an analytical balance with motorized doors, an HPLC system, and a mobile liquid handler. We created a declarative specification (in YAML) of the lab for EOS containing 8 device definitions and 17 task definitions. These devices and tasks included the relevant ones from the LLE platform as well as extraneous devices (centrifuge, hot plate, pH meter, UV-Vis spectrometer) and extraneous tasks (e.g., ``Centrifuge Sample,'' ``Measure pH,'' ``Heat And Stir''). We added the extraneous devices and tasks as distractors to test whether the AI agent can identify the correct subset of building blocks and ignore irrelevant options when creating a protocol.

We prompted the agent with: ``Weigh the HPLC vial at slot A1 in lle\_lab.'' Across 10 trials, the agent produced semantically correct protocols with a 100\% success rate, consistently selecting only the relevant devices and tasks and ignoring all distractors. Table~\ref{tab:agent_metrics_lle} reports execution metrics.

\begin{table}[t]
    \vspace{0.5em}
	\caption{AI agent execution metrics for LLE vial weighing protocol generation (N=10 trials).}
	\label{tab:agent_metrics_lle}
	\centering
	\renewcommand{\arraystretch}{1.1}
	\begin{tabular}{@{}lcc@{}}
		\toprule
		\textbf{Metric} & \textbf{Mean $\pm$ Std} & \textbf{Range} \\
		\midrule
		Wall Time (s) & 71 $\pm$ 13 & 57--106 \\
		Reasoning Steps & 3.0 $\pm$ 0.0 & 3--3 \\
		MCP Tool Calls & 2.0 $\pm$ 0.0 & 2--2 \\
		Validation Corrections & 0.0 $\pm$ 0.0 & 0--0 \\
		LLM Cost (USD) & 0.20 $\pm$ 0.01 & 0.18--0.22 \\
		\bottomrule
	\end{tabular}
\end{table}

\begin{figure}[t]
    \vspace{0.5em}
	\centering
	\includegraphics[width=\linewidth]{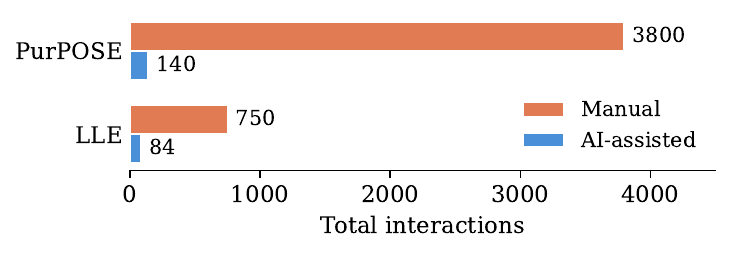}
	\caption{Approximate number of minimum discrete interface actions required to fully specify each protocol through manual authoring versus AI-assisted authoring.}
	\label{fig:interaction_comparison}
\end{figure}

\subsection{Impact of the AI Agent on Interaction Complexity}
To quantify the reduction in interaction complexity afforded by the AI agent, we measured the minimum number of discrete interface actions (mouse clicks and keystrokes) required to fully specify each protocol through manual authoring versus AI-assisted authoring. The minimum number of interface actions is the lowest number of actions one would need to take to implement the protocol in the visual protocol graph editor or to prompt the AI agent and answer its questions, without making any errors. We computed the number of actions for manual editing by calculating the number of clicks required to create task nodes, to select fields such as input parameter fields to edit them, and to type any values. As shown in Figure~\ref{fig:interaction_comparison}, the PurPOSE crystallization protocol required approximately 3800 manual actions versus 140 AI-assisted actions (27$\times$ reduction), and the LLE vial weighing protocol required approximately 750 manual actions versus 84 AI-assisted actions (9$\times$ reduction).

\subsection{Failure Analysis and Limitations}

The primary source of errors was the AI agent making assumptions rather than asking the scientist. For example, in the LLE protocol the agent assumed the balance did not require zeroing and in the color mixing campaign the agent arbitrarily made mixing parameters continuous or discrete. These assumptions produce structurally valid but semantically imprecise protocols. The biggest area for improvement is enabling the AI agent to better deal with ambiguity, whether by actively seeking clarifications or via parallel generation of candidate protocols with variance analysis. We also encountered one instance where the AI agent generated incorrect Python code for the optimizer; our validation layer covers only YAML, but extending it to parse and compile Python would address this gap. Additionally, the AI agent's input context can be optimized: providing full specifications for all tasks and devices is unnecessary when only a subset is relevant, and a two-stage approach that first provides names and descriptions and then retrieves full specifications on demand would reduce token usage and inference costs.


\section{Conclusions}

We presented an AI agent architecture that integrates large language models with laboratory orchestration, enabling scientists to interactively create and monitor automated lab protocols using natural language. Integrated into the Experiment Orchestration System (EOS), the AI agent supports the complete experiment lifecycle, including creating protocols, running optimization campaigns, and analyzing experimental data. A visual protocol graph editor enables seamless alternation between AI-assisted and manual protocol construction. Evaluations demonstrated 97\% first-attempt protocol generation success on a simulated color mixing campaign across 65 trials on four different problems, and correct creation of multi-phase protocols and multi-objective optimization campaigns for two self-driving chemistry platforms from the literature, with an order-of-magnitude reduction in interaction complexity. Our evaluation focused on functional correctness and interaction complexity; a formal usability study with laboratory scientists across diverse domains remains important future work.


\section*{Acknowledgement}

Figure~\ref{fig:header} was created with the aid of Google's Nano Banana 2 generative AI model.


\bibliographystyle{IEEEtran}
\bibliography{references}

\addtolength{\textheight}{-12cm}

\end{document}